\documentclass[sigconf]{acmart}

\usepackage{booktabs} 

\usepackage[utf8]{inputenc}
\usepackage{slashbox}

\setcopyright{rightsretained}

\usepackage{flushend}

\acmConference[KDD Workshop on AI for Fashion]{KDD}{2018}{London, UK}

\usepackage{soul}
\usepackage{color}

\begin{document}

\title{A Unified Model with Structured Output for Fashion Images Classification}


\author{Beatriz Quintino Ferreira}
\affiliation{%
 \institution{ISR, Instituto Superior T\'{e}cnico, Universidade de Lisboa, Portugal}
}
\email{beatrizquintino@isr.tecnico.ulisboa.pt}

\author{Luís Baía}
\affiliation{%
 \institution{Farfetch}
}
\email{luis.baia@farfech.com}

\author{João Faria}
\affiliation{%
 \institution{Farfetch}
}
\email{joao.faria@farfetch.com}

\author{Ricardo Gamelas Sousa}
\orcid{0000-0001-8822-5412}
\affiliation{%
 \institution{Farfetch}
}
\email{ricardo.sousa@farfetch.com}

\renewcommand{\shortauthors}{Beatriz Quintino Ferreira et. al}

\begin{abstract}
A picture is worth a thousand words. Albeit a cliché, for the fashion industry, an image of a clothing piece allows one to perceive its category (e.g., dress), sub-category (e.g., day dress) and properties (e.g., white colour with floral patterns). The seasonal nature of the fashion industry creates a highly dynamic and creative domain with evermore data, making it unpractical to manually describe a large set of images (of products). 


In this paper, we explore the concept of visual recognition for fashion images through an end-to-end architecture embedding the hierarchical nature of the annotations directly into the model. Towards that goal, and inspired by the work of~\citep{Hu2016}, we have modified and adapted the original architecture proposal. Namely, we have removed the message passing layer symmetry to cope with Farfetch category tree, added extra layers for hierarchy level specificity, and moved the message passing layer into an enriched latent space.

We compare the proposed unified architecture against state-of-the-art models and demonstrate the performance advantage of our model for structured multi-level categorization on a dataset of about 350k fashion product images.

\end{abstract}

%
%



\maketitle


\section{Introduction}
\label{sec:Intro}
\begin{figure}[!t]
\includegraphics[width=0.50\textwidth]{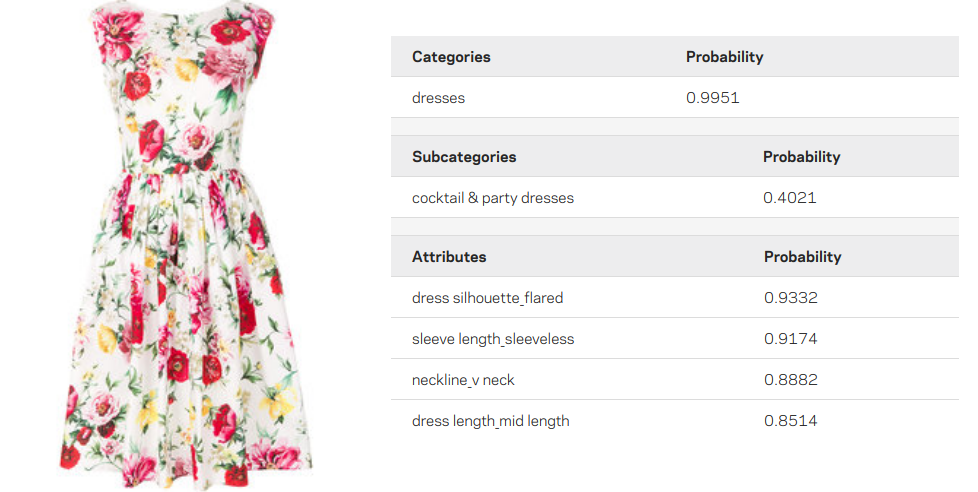}
\caption{A pictorial example of the result of our automatic multi-level categorization system.}
\label{result_example}
\end{figure}
Image classification is a classical Computer Vision problem. Although the advent of Deep Convolutional Neural Networks gave a dramatic push forward, there is an increasing interest to describe images and its properties in a richer way. In this vein, more attention has been drawn to multi-label problems and also to the exploration of label relations. 


Such rich descriptions are fundamental in several e-commerce businesses. Since it is inherently a very visual domain, specifically for items search, exploring visual information for image categorisation is key to enhancing product exploration and retrieval. In fact, in these businesses, it is crucial to retrieve relevant images (with the desired characteristics) from the users' textual queries. Moreover, considering the continuous stream of new products being added to the catalogues, the automatic generation of image labels can alleviate the workload of human annotators, improve quick access to relevant products, and help generate textual product descriptions. 

Specifically, we take advantage of \emph{a priori} structural knowledge together with the rich relational information that exists among product labels and associated concept semantics to improve image classification, thus enhancing product categorisation and consequent retrieval. We will focus on a real use case that can be easily generalised for any e-commerce platform. Namely, at Farfetch, a key player in the luxury fashion market, we aim to improve product categorisation and attribute prediction exclusively from visual features. 

The current Farfetch category tree has five levels (gender, family, category, sub-category, and attributes), where all levels but one (attributes level) are mutually exclusive. Thus, the challenge we propose to address is to simultaneously estimating class predictions for all levels of the category tree for an image, while explicitly exploring the structure provided by the mentioned semantic hierarchy (Figure~\ref{result_example}). Our motivation is to reduce the number of specialized classification models to train, without compromising (or even, preferably, improving) the performance.

Furthermore, for multi-label classification at the attribute level, we face a large label space with imbalanced distribution and variable length prediction, which hinders traditional models.

The contributions of this work are as follows:
\begin{itemize}
\item A new method to categorise and predict attributes of fashion items exclusively from visual features;
\item The proposed method is a unified end-to-end deep model that jointly predicts different concept levels from a hierarchy tree, thus incorporating the concepts structure;
\item We show experimentally that the unified approach outperforms state-of-the-art models specialised for each concept level.
\end{itemize}

The remainder of this paper is organised as follows. In Section~\ref{sec:RelWork} we briefly review related works. In Section~\ref{sec:method} we introduce and provide details of our baseline and final model for structured output image classification. Section~\ref{sec:results} presents and discusses the experimental results. Finally, in Section~\ref{sec:conclusion} we draw the conclusions.

\section{Related Work}
\label{sec:RelWork}


Powered by the creation of large-scale annotated image datasets such as the ImageNet~\cite{imagenet_cvpr09}, deep convolutional neural networks (CNNs) are known to produce state-of-the-art performance on many visual recognition tasks. They have been effectively used as universal feature extractors, either in an “off-the-shelf” manner or through a small amount of “fine tuning”. Among these currently called "standard" CNNs we can find the VGG~\cite{Simonyan2014}, ResNet~\cite{He2016}, or the InceptionNet~\cite{Szegedy2016}.

Many early works approach the multi-label problem as an extension of the single label methods, learning an independent classifier for each one. 
The previous approaches, nevertheless, do not capture label relations nor impose any \emph{a priori} known structure. 
More recent approaches extend traditional CNNs and learn to incorporate label relations. 
One of the first works to explore and enforce structure on object classification applying deep neural networks was~\cite{Deng2014}. This work introduces the Hierarchy and Exclusion Graphs formalism that enables encoding relations between labels, thus exploring the rich structure of real world label semantics, rather than considering all labels as independent and lying in a flat structure. On top of the previous formalism, this work proposes an inference algorithm, from the Conditional Random Field family, that uses label relations as pairwise potentials. The algorithm is implemented as a standalone layer in a deep neural network that can be added to any feed-forward architecture.
Nevertheless, since their focus is mainly on the definitions and theorems supporting the proposed formalism and not on specific architectures to implement it, the contribution of~\cite{Deng2014} is primarily at a theoretical level, and thus less pertinent to our final application.  
%
%
%
%
%
%

The work in~\cite{Yang2015} also addresses the problem of using structured (e.g. hierarchical) prior knowledge of the image classes and labels to aid classification. However, it follows a different approach as the multi-level \emph{a priori} reasoning is achieved by directly performing alterations to the deep neural network architecture. This work revisits multi-scale CNNs to propose a new network architecture structured as a directed acyclic graph that feeds the several multi-scale features to the output layer.

In the same vein, as labels are not semantically independent, the work in~\cite{Hu2016} takes advantage of label relations to enhance image classification. The proposed model takes as input an image and the graph of label relations which encodes the underlying hierarchical semantic relations. This way, the proposed deep neural network architecture allows to encode both inter-level (hierarchical) and intra-level label relations. This is shown to improve inference over layered visual concepts. The application example given in~\cite{Hu2016} takes the WordNet taxonomy as external knowledge, expressing it as a label relation graph, and learns the structured labels.
Resorting to this framework facilitates the information passing (instantiated by the message propagation algorithm proposed in this paper) in the deep network.
It is worth noting that this multi-level structured prediction problem can be interpreted as an instance of multi-task learning, since the latter outputs estimates for multiple (different but related) tasks. For more on the multi-task take of the structured learning proposed in~\cite{Hu2016} please refer to~\cite{Long2017}.
The introduced message passing scheme for structured semantic propagation is built on top of a state-of-the-art deep learning platform, in this case a CNN. To predict the outputs for each level the method adds a final loss layer.
The first method proposed in~\cite{Hu2016} is the Bidirectional Inference Neural Network (BINN), a Recurrent Neural Network (RNN)-like algorithm that integrates structured information for prediction. BINN architecture captures both intra-level and inter-level relations through two model parameters, one capturing two-way label relations between two consecutive concept levels, and the other accounting for the label relations within each concept level.

Subsequently, both~\cite{Liu2017} and~\cite{Niu2017} follow approaches similar to~\citet{Hu2016} to integrate structure in label prediction. However, the first tackles not only the multi-label classification, but also the captioning problem (to do so, this work applies the CNN-RNN encoder/decoder design pattern, similar to~\cite{Wang2016}, which has become popular to address structured label prediction tasks); while the second performs multi-modal feature learning by concatenating the visual and textual (extracted from social tags) feature vectors. Furthermore,~\cite{Niu2017} assumes training images, ground-truth class labels, and noisy tags are available as training input, and also that test images with respective noisy tags are available at test time.
The above assumptions, though, do not occur in our case, as we do not have noisy tags for our test images.
Therefore, although these works seem very interesting and report very promising results (potentially outperforming the method from~\cite{Hu2016}), we found them not to be applicable to our case, and thus beyond the scope of this paper.

Publicly available datasets include the Clothing Attributes Dataset \cite{ClothingAttDataset2012}, Fashionista~\cite{Fashionista2012} and DeepFashion, a large scale clothing dataset published in~\cite{LiuDeepFashion2016}. The previous datasets are a reference in the fashion scope and can be used for several machine learning tasks, with the most prominent one to our work being the classification of categories and attributes of fashion products. However, all these datasets lack a hierarchical structure of the annotations, in the sense that there is not a defined hierarchy between attributes and categories, nor among the categories itself. Given that a core part of our contribution relates to the benefit of embedding a hierarchy into the model framework exploring label relations, we claim that none of these datasets are suitable for our showcase, and thus have decided to not use them in our experiments.
%


Focusing on approaches with application on labelling of fashion items, we refer the works~\cite{Sun2015, Corbiere2017, Laenen2017}.
Particularly,~\cite{Sun2015} performs clothing style and attribute recognition via training specific detectors (with traditional hand-crafted features) for each fashion attribute.
~\cite{Corbiere2017} crawled the internet to gather a dataset from fashion e-commerce websites to perform weakly supervised image retrieval and tagging. This work trains two different and independent deep models to perform multi-class categorisation and attribute labelling.
In~\cite{Laenen2017} a deep model is also trained on a fashion dataset to perform image retrieval. The goal of the latter work is cross-modal search (using text as an additional source of information), and rule-based image operations are applied to the dataset.  
Yet, none of the previous works account label relations or hierarchical structure in a unified model, thus differing from our proposed method. 

%
%
%
%

\section{Methodology}
\label{sec:method}
\subsection{The task}
\begin{figure}[!t]
\includegraphics[width=0.4\textwidth]{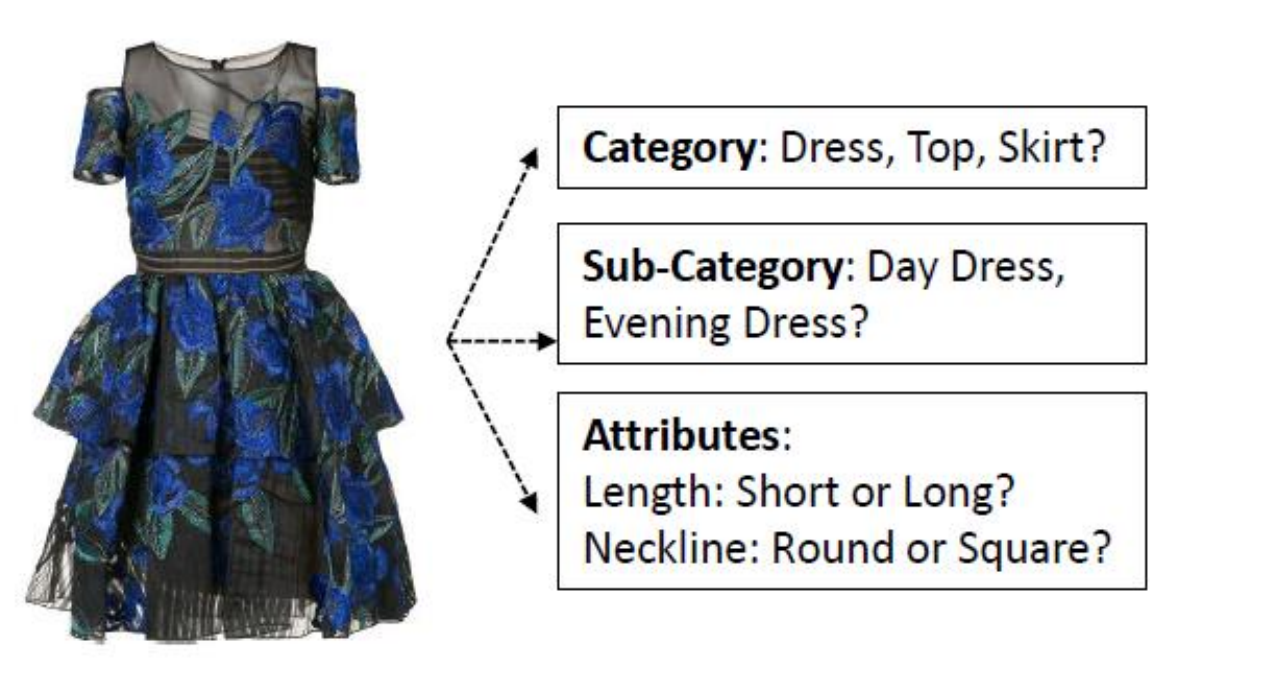}
%
%
\caption{Each image is associated with visual concepts from several levels. We aim to jointly predict the classes/attributes for all levels only using visual features.}
\label{fig:concept_level_class}
\end{figure}

Our goal is to classify an input image across our entire categorisation tree (see Figure~\ref{fig:concept_level_class}).
The current Farfetch category tree is composed by five categorisation levels: gender, family, category, sub-category and attribute. More specifically, the family, category, and sub-category levels are mutually exclusive, while at the attribute level, a product can have more than one attribute. As a consequence, when we approach the problem by jointly classifying all levels of a product (using its image), the classifications at the category and sub-category levels are multi-class classification problems, whereas at the attribute level we face a multi-label classification problem. The family, category, and sub-category levels follow a hierarchical structure, meaning that knowing a child level allows the parent level to be directly inferred. On the other hand, there are some attributes shared by different categories, preventing a direct inference of the parent level.

The family and gender are not considered here, while the remaining three levels will be learnt. The reason to discard the former two relates to the fact that the family level is so generic that without a proper category level prediction provides little value. On the contrary, most categories and sub-categories are gender-specific, thus the estimated category and sub-category automatically reveal the gender.

In summary, our task is defined as follows: given the image of a product, classify its proper category, sub-category and attributes (as depicted in Figure~\ref{fig:concept_level_class}). 
The gender and family are automatically inferred from the respective predictions.

\subsection{First Approach}
\label{subsect:first_approach}
For well defined and classic visual classification problems, with a single classification output, the current state-of-the-art approaches have proved to be quite capable, as previously discussed in Sections~\ref{sec:Intro} and~\ref{sec:RelWork}. 

By decomposing the previously mentioned task as three independent categorisation problems, we can directly plug in most of the standard state-of-the-art models to develop the first approach. Namely, a custom ResNet-50~\cite{He2016} fine-tuned for our domain can be instantiated to solve each learning task individually. The proposed base architecture for our first approach is a ResNet-50 connected to a Multilayer Perceptron of size 1024 with a ReLU function as the activation layer followed by another Multilayer Perceptron in the output dimension space. Depending on the type of output (multi-class or multi-label), the final activation function will be a softmax or a sigmoid function, respectively. The Inception~\cite{Szegedy2016} and VGG~\cite{Simonyan2014} were also considered as the base CNN, however the ResNet-50 achieved slightly better performance.

\begin{figure*}[!t]
\includegraphics[width=0.8\textwidth]{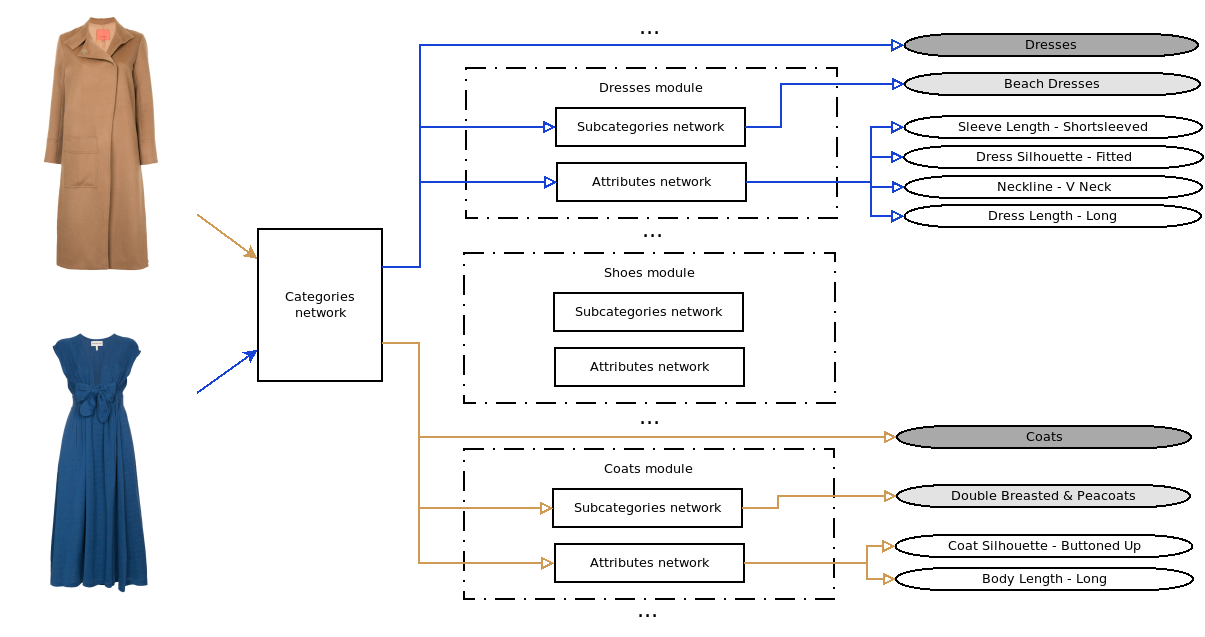}
\caption{A schematic representation of our first approach architecture.}
\label{fig:original_approach}
\end{figure*}

Considering this model architecture as the "template", we implemented a pipeline of such specialised deep "template" neural network models as our baseline (see Figure~\ref{fig:original_approach}). More precisely, the first model predicts the category for each product and then, depending on the estimated category (e.g. dress category), the image is fed to a second model, specialised in the sub-categories of that specific predicted category (e.g., dress type sub-categories). The same reasoning is applied to attribute prediction, i.e., a specific type of attribute predicting model (e.g. specific model for dress length or dress silhouette type of attributes) is invoked if the product is firstly predicted to belong to the category dress. This approach allows the creation of models specialised into very small tasks that could potentially achieve good results when combined in a large group of "simple models". However, this forms a pipeline of specific models dependencies that is not taking advantage of the underlying level relations/structure, and also is not scalable to cover all product types, as numerous models would have to be trained and maintained.

Regarding the training process, only the final eleven layers of the ResNet-50 (corresponding to the last convolutional block) are allowed to have its weights re-trained. The chosen optimiser was Adam~\cite{Kingma2015}. The choice of the complete architecture, depicted in Figure~\ref{fig:original_approach}, was the result of a comprehensive process of experimentation supported by the most recent findings in the literature.

\subsection{Proposed model}
\label{subsect:proposed_model}
\begin{figure}[!t]
\includegraphics[width=0.4\textwidth]{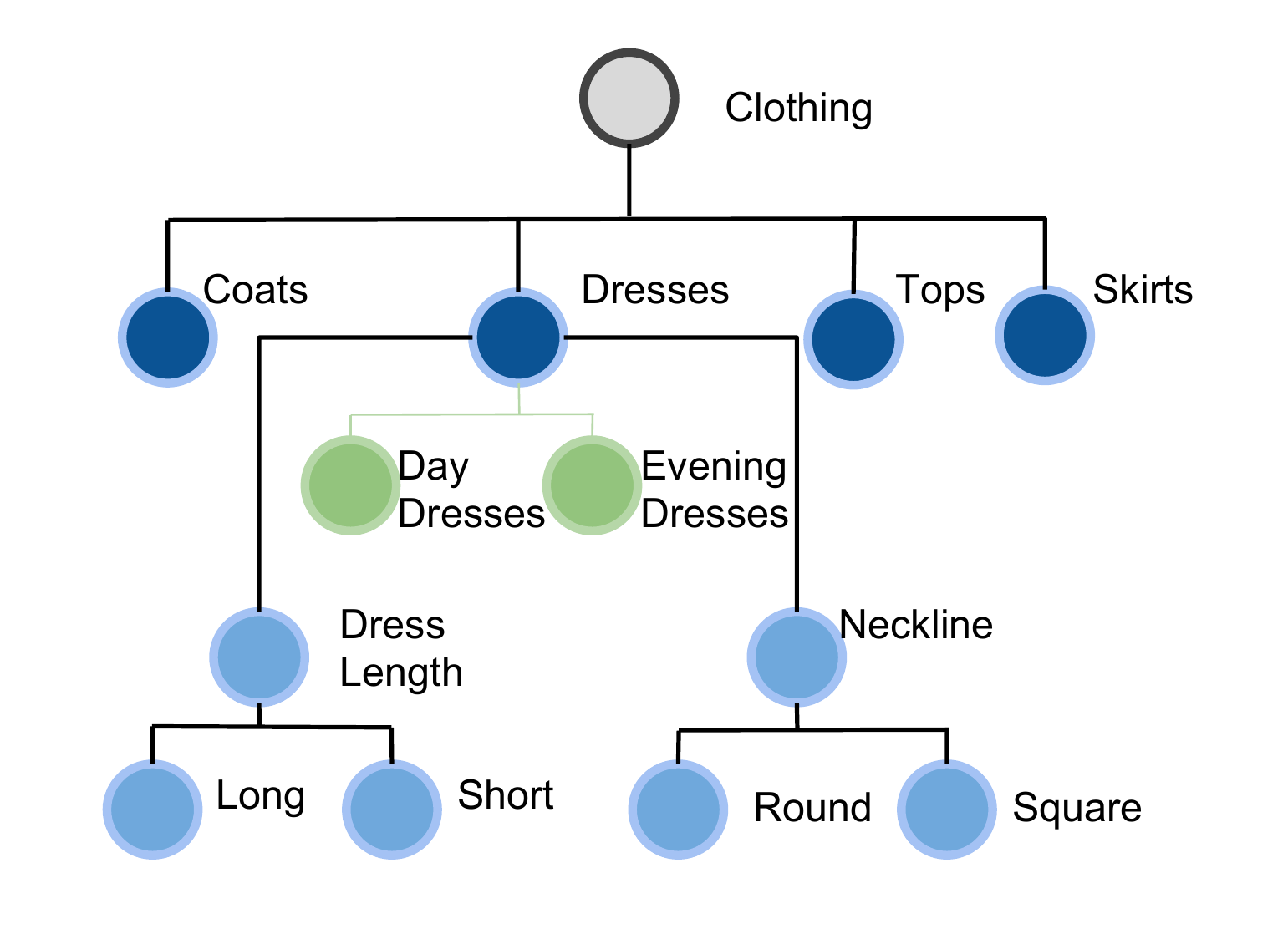}
\caption{Category tree example. Visual concepts are hierarchically structured and thus we can use graphs to represent different level concepts (categories, sub-categories and attributes) relations.}
\label{fig:category_tree}
\end{figure}

Let us consider an example of the intuition behind the proposed structured learning in the Farfetch scenario (see a category tree example in Figure~\ref{fig:category_tree}): we know that day dress (sub-category) is a dress (category), t-shirts and jerseys (sub-category) are tops (category), and since the dresses and tops categories are mutually exclusive, an image with a high score for dress should increase the probability of the (sub-category) day dress while decreasing the probability for (sub-category) t-shirt and jerseys. The same reasoning can be applied seamlessly for the relation between categories and attributes.

\begin{figure*}[!t]
\includegraphics[width=0.95\textwidth]{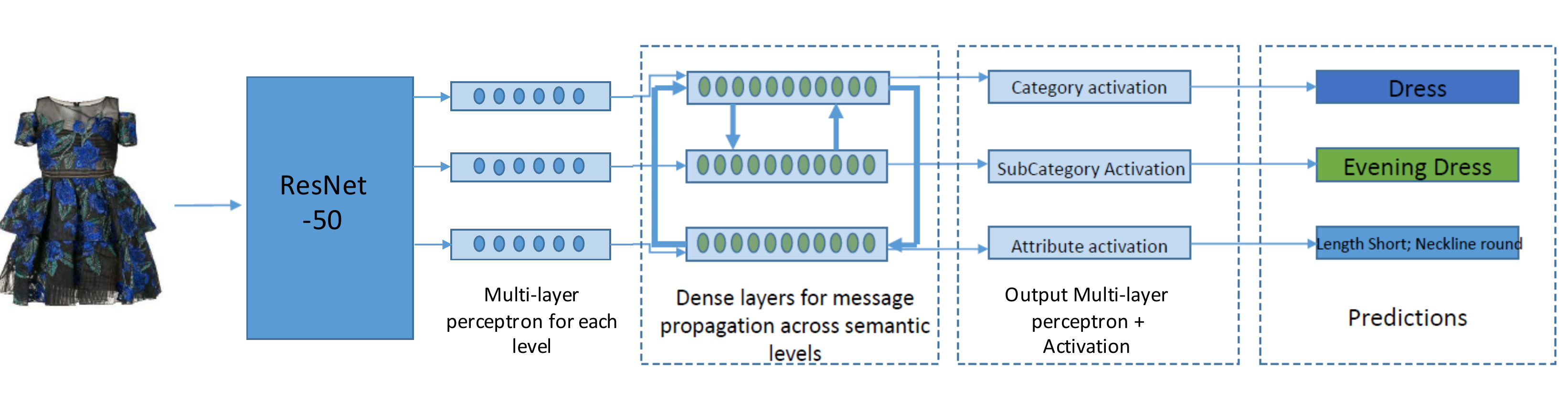}
\caption{Proposed unified model high-level architecture: category, sub-category and label prediction framework.}
\label{fig:final_architecture}
\end{figure*}

\begin{figure*}[!t]
\includegraphics[width=0.95\textwidth]{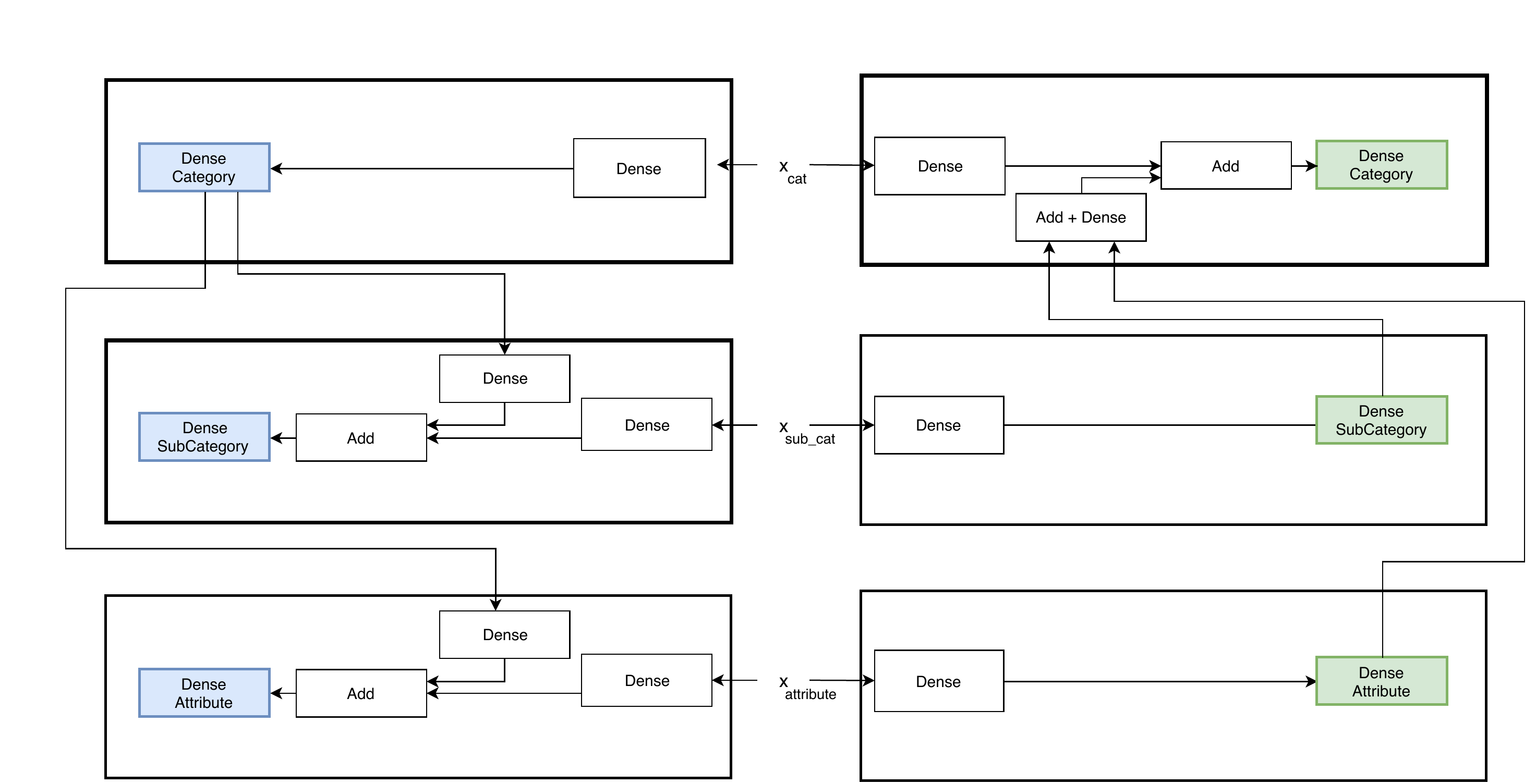}
\caption{First part of the Message Passing block (downwards and upwards inter-level and intra-level propagation). }
\label{fig:mp_part1}
\end{figure*}

Our proposed architecture (shown in Figure~\ref{fig:final_architecture}) is inspired by the BINN method of Hu et al.~\cite{Hu2016}. However, unlike~\cite{Hu2016} who assumes symmetry in the message propagation, we adapted this architecture to the Farfetch scenario, where the category level influences the sub-category (and vice-versa), and also the attribute level (and vice versa). Yet, note that there is no direct influence between the sub-category and attribute levels (see Figure~\ref{fig:category_tree}), and our message propagation scheme is not symmetrical with respect to the concept levels (see also message propagation block of Figure~\ref{fig:final_architecture} and Figure~\ref{fig:mp_part1}). Although such influence, between the subcategory and attribute levels, might be logical, we opted to stick with the structure that reflects our business model category tree, not to mention that including this influence would considerably increase the model complexity. Moreover, instead of applying the Message Propagation in the output space, we have decided to do it in the latent space of higher dimension. The intuition relies on the concept that propagating each level distinctive features (thus using the latent space) to the other levels should enhance the results even further than propagating just a mere certainty or doubt (output dimension space) regarding a classification per level.

After performing some preliminary tests with the previous architecture we saw indications that the latent image feature vector output by the ResNet~\cite{He2016} was not able to generalise well to the three (with very different specificities) concept levels. Therefore, to fix this symptom, we have tried two variants of the architecture: a first one where a Multilayer Perceptron is added for each level between the ResNet output and the beginning of the Message Propagation block, allowing the CNN output to be modified into an enriched and specialised dimensional space before feeding the Message Propagation block; and a second approach that tries to embed this specificity information into the ResNet itself. Specifically, we individually retrain the final layers of ResNet~\cite{He2016} for each level, thus increasing the number of layers inside our CNN box.

As we will discuss in the Experimental Results section, the first approach outperforms the latter.

\begin{figure}[!t]
\includegraphics[width=0.45\textwidth]{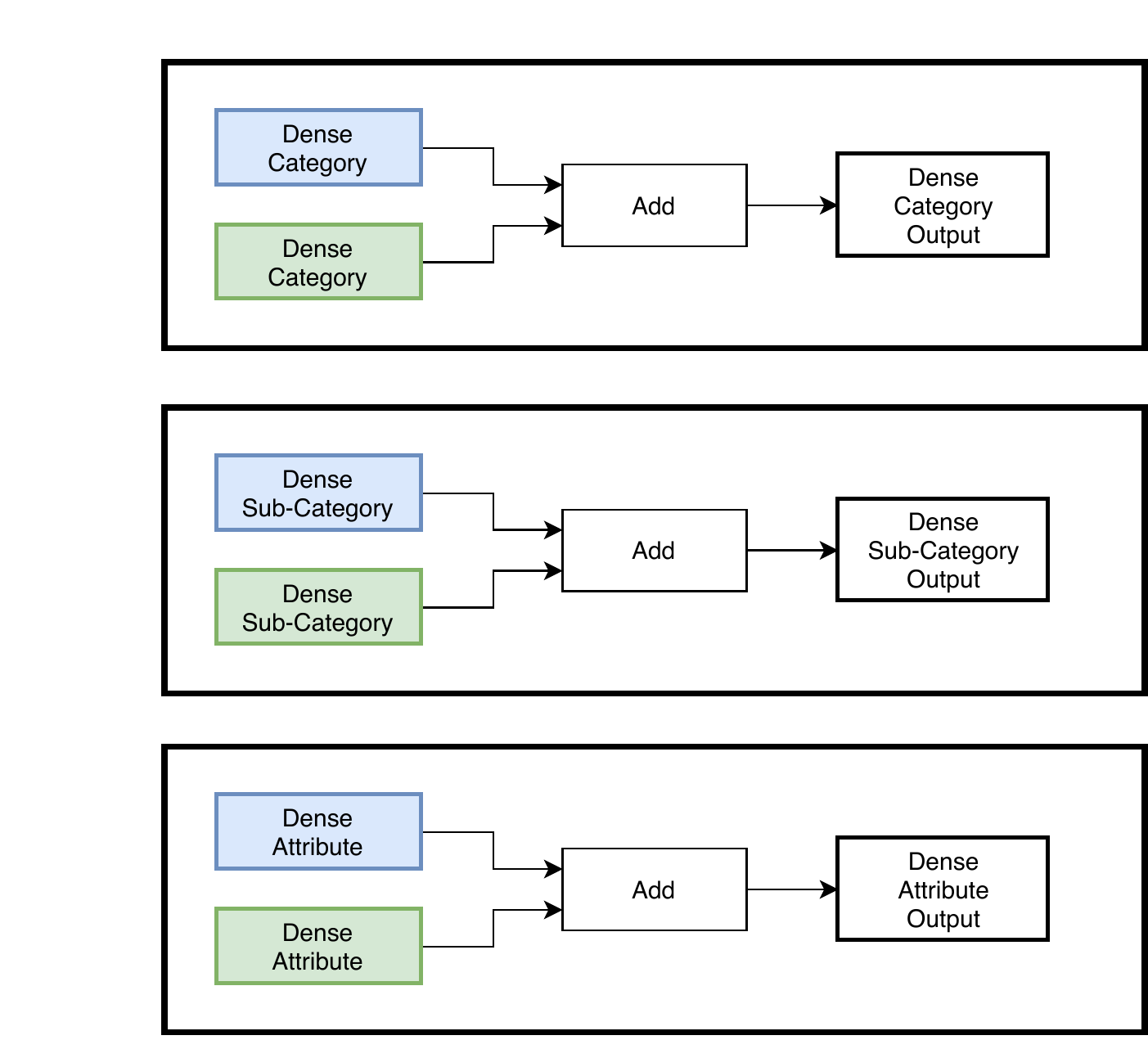}
\caption{Second part of the Message Passing block (merging the result of each direction from the first part of the message passing).}
\label{fig:mp_part2}
\end{figure}

\paragraph{\textbf{Message Propagation Block}}
\label{subsect:message_propagation}
The message propagation block is ultimately responsible for the hierarchical structure of our proposed model. Since our implementation differs in some aspects from the original one from~\cite{Hu2016}, we present it in more detail. Figures \ref{fig:mp_part1} and \ref{fig:mp_part2} show the architecture of our message propagation block.

This block has three inputs, a latent vector per level: $x_{cat}$, $x_{sub-cat}$, $x_{attribute}$ standing for category, sub-category, and attribute, respectively. The architecture resembles a traditional bi-directional layer with each direction being shown on the left and right side of Figure \ref{fig:mp_part1}. The choice of a bi-directional approach to instantiate the intra-level relations is supported by the idea that the information from relevant patterns to detect a category (category latent information) could potentially benefit the extraction of relevant patterns towards a sub-category prediction, so would the other way around.

On the left side of Figure \ref{fig:mp_part1}, we start on the highest level of the hierarchy and propagate the category latent information towards the sub-category and attributes levels. On the right side, the information is propagated on the inverse direction. Notice that there is no direct connection between the sub-category and attribute level. This connection is not present in our hierarchy tree, and adding that connection to the network would considerably increase the number of parameters to be learned without any strong reason to support such an increase in complexity.

The levels that are lacking a Dense and Add layers (see Figure~\ref{fig:mp_part1}) are explained by the fact that they are not receiving any information from another conceptual level from the hierarchy. Namely, on the top to bottom direction, this happens on the category level, since it is the top level that propagates to the sub-category and attribute level. On the bottom to top direction, both the sub-category and attribute levels are simpler, since both are the source of upwards propagation to the category level.

The boxes in blue and green in Figure \ref{fig:mp_part1} are the intermediate outputs of each direction of the message passing block per hierarchy level. Each pair has still to be merged into a single vector. Figure \ref{fig:mp_part2} shows how this last step is achieved. Essentially, each pair is summed with an extra Dense layer. The three outputs of the Message Propagation are then fed into the Output Multi-Layer Perceptron.

\paragraph{\textbf{Implementation Details}}
We implemented our proposed model in Keras, with TensorFlow as the backend.
The message passing block (that encodes the category tree) is built on top of the off-the-shelf convolutional neural network ResNet-50~\cite{He2016}, pre-trained (i.e., with weights initialised as the weights learned after training the network) on the ImageNet~\cite{imagenet_cvpr09}. Additionally, three parallel dense layers (one per hierarchy level) of dimension 1024 are connected to the output of the ResNet-50. These will be the inputs of the Message Propagation block.

Every Dense layer defined in the Message Propagation block is of dimension 1024 with a L2-norm regularization and regularization factor of $0.0005$ (promoting the learning of more uniform weights, thus reducing the risk of over-fitting) followed by ReLU activation layers. The final Dense layers of this block (the Dense layers shown in Figure \ref{fig:mp_part2}) are also followed by a Dropout~\cite{Hinton_droput2012} of rate $0.3$.

The full architecture (encompassing the ResNet-50, intermediate dense layers for each level and the message passing block) totals $46.915.690$ trainable parameters.

Output activations for each level predictions depend on the problem at hands, i.e., as the category and sub-category level predictions are multi-class problems we use a softmax function as activation, while at the attribute level we have a multi-label problem and thus we use a sigmoid activation function.

The network is trained by minimising a weighted cross-entropy loss for each level in order to estimate the parameters that originate the most correct predictions for the category, sub-category and attribute levels. A weighting mechanism is used to address class imbalance, a common issue that also arises in our dataset. In particular, we compute the occurrence frequency of each class/label and apply a customised cross-entropy loss where the penalisation is weighted by the inverse of its frequency. Hence, the loss for predicting more frequent classes is down-weighted while when predicting more rare classes the loss is penalised. This way, all classes per level should be equally important during the training process of the model. 
Also, contrarily to what is presented in~\cite{Hu2016}, we train our model in a single-shot fashion (end-to-end). The loss functions are optimised via backpropagation and batched-based Adam~\cite{Kingma2015}, with a batch size of 32 images and a learning rate of 0.001 for this optimiser.
Although stochastic gradient methods may not converge to an optimal point, as discussed in~\cite{Reddi2018}, Adam and its variants have empirically demonstrated leading performance when compared to other state-of-the-art optimisers~\cite{Reddi2018}.

During the training phase, to perform data augmentation, we apply random transformations (including flipping, cropping and rotating) to the input images (with each input having a fixed probability of 50\% of suffering such a transformation). Augmenting the dataset with synthetically generated images from the input images has been proved to be a particularly effective way of regularizing deep neural networks~\cite{Simard2003,Krizhevsky2012}. Specifically, we have seen great performance improvements after including data augmentation in our model (a further analysis of these results, though, is out of the scope of this work).

\section{Experimental Results}
\label{sec:results}
\subsection{Dataset}
\label{sec:Dataset}

Our dataset is composed of Farfetch front-side images, with a single product centred on a homogeneous and clean white background, and the associated ground truth labels for each concept level (one category label, one sub-category label, and, potentially, one or multiple labels for the attributes level). It is important to highlight that data is manually annotated and may not always be consistent, namely at the attributes level. For example, one t-shirt may have sleeve length and neckline shape annotations while a different t-shirt may only be annotated relative to sleeve length (i.e. missing data). This brings extra difficulty to our problem.

The assembled dataset contains 356,553 products and their corresponding images (one per product) of size 255x340 (these images are then resized to the ResNet input dimension of 224x224 pixels). As stated, each product has an associated category and sub-category, according to the categories and sub-categories from Farfetch category tree. The former consists of 64 possible labels, while the latter has 95 possible values. As expected, this is a highly unbalanced dataset. Each category has on average 5571 products, the most frequent one (tops) having 42565 products, and the less numerous (fine bracelets) containing only 307 products. On the other hand, for sub-categories the mean count is 2306, the maximum (for jumpers) is 15523, and the minimum (for sport tank tops) 250. In terms of attributes, there are 75 of them. The most frequent one (occasion/casual) appears in 38624 products, and the less common (neckline/halterneck) shows up in 330 products. The mean value is 4573 products per attribute. Conversely, each product has, on average, 0.96 attributes. The number of attributes per product ranges between 0 and 5. Refer to Tables~\ref{tab:entity_cardinality_table} and~\ref{tab:products_count_table} for a concise description of our dataset. We consider a 75\%/25\% train/test split.
%
%

\begin{table}[!t]
\centering
\begin{tabular}{ l c }
\toprule
 & \textbf{cardinality} \\ \midrule
products & 356553 \\ 
categories & 64 \\ 
sub-categories & 95 \\ 
attributes & 75 \\
\toprule
\end{tabular}
\caption{Assembled dataset entity cardinality.}
\label{tab:entity_cardinality_table}
\end{table}

\begin{table}[!t]
\centering
\begin{tabular}{ l c c c }
\toprule
 & \textbf{mean} & \textbf{max} &\textbf{min} \\ \midrule
products / category & 5571 & 42565 & 307 \\ 
products /sub-category & 2306 & 15523 &  250 \\ 
products /attribute & 4574 & 38624 & 330 \\
attributes / product & 0.96 & 5 & 0 \\
\toprule
\end{tabular}
\caption{Average, maximum and minimum number of products per hierarchical level and attributes per product.}
\label{tab:products_count_table}
\end{table}

\subsubsection*{\textbf{Evaluation Metrics:}}
For category and sub-category classification we choose the class with highest estimated confidence score. For the multi-label attribute classification, the labels are predicted as positive if the predicted label confidence is greater than 0.75. This threshold was chosen in conjunction with the business (to find a good ratio between adding new attributes without making serious mistakes). Nevertheless, we will present some metrics that are threshold independent to allow a better comparison between each approach.

For the multi-class problems, for category and sub-category levels, we report overall precision (OP), recall (OR), and F1-score (OF1), weighted by class support, i.e., the number of true instances for each class. Therefore, given that we are using a weighted version of the macro precision and recall, the resulting F1-scores may not be between precision and recall. 

For the multi-label classification, at attributes level, we also employ overall precision (OP), recall (OR), and F1-score (OF1) for performance comparison. Moreover, we use precision (P$@$k), recall (R$@$k) and F1-score (F1$@$k) $@$ top K labels (where K is the number of ground truth labels that each product is annotated with). We also report the average precision (AP), which summarises the precision-recall curve. The previous metrics allow us to assess the method performance irrespective of defining a threshold on the confidence scores for positive/negative classification. 
%
%
For all these metrics, the larger value, the better the performance.
Specially for the fashion e-commerce business, a high recall for attribute labelling is very relevant in terms of platform usability, since the more attributes that are associated to a product, the more products can be found/indexed by a larger set of filters/queries, thus improving product discoverability. 
Moreover, since the ground truth annotations naturally contain missing labels (products of the same category may not be all annotated for the same visual properties), recall $@$ top K turns out to be a rather strict evaluation metric. Therefore, we also present qualitative results to assess our model's performance (see Figure~\ref{fig:qual_results}).

\subsubsection*{\textbf{Compared methods:}}
To validate and evaluate the performance of the proposed method we compare it against our first approach (pipeline of different ResNet-50 for each category level, see Section~\ref{subsect:first_approach} and Figure~\ref{fig:original_approach}), which we dub as \emph{baseline}. We also design some variations of our proposed method (see Section~\ref{subsect:proposed_model}), to further validate its effectiveness as a whole and of some of its parts (performing a brief ablation analysis). Specifically, the first variation is equal to our model but with the final 11 layers (equivalent to the last convolutional block (\textit{block 5c})) of ResNet-50 independently re-trained for each output level, while the second variation is our proposed model with the message passing scheme replaced by dense layers, independent from each other, after each ResNet-50 output. We designed the latter alternative so we could assess the importance of the Message Passing in our network. However, simply removing it from the network would not lead to a clear and fair comparison given that the final number of parameters  would be severely reduced. Therefore, we have replaced the Message Passing block by consecutive Dense layers of dimension 1024 per level so that, approximately, the number of trainable parameters would remain the same as in our proposed model.

With the first variation, denoted as \emph{Ours - ResNet Indep}, we investigate whether trying to obtain independent high-level features (from ResNet-50), that will be possibly more specific and discriminative for each different level, can be beneficial to our problem. Whereas with the second alternative, referred as \emph{Ours - No MP}, we aim to validate the inclusion of the block of message passing among our concept levels.

\subsection{Results and Discussion}

\subsubsection*{Category level}

Experimental results on our dataset concerning category level are shown in Table~\ref{tab:results_cat_2}. To obtain these results we used all images from our dataset and, as ground truth, the category annotations to train the baseline method, and the annotations from all levels to train the variants of our proposed model. We can observe that all variants of the proposed method outperform the baseline, thus supporting our unified approach. In particular, our final approach attains better results than its variants for all considered metrics. This seems to indicate that both sharing the final ResNet-50 layers and including the message passing scheme inspired in~\cite{Hu2016} are valuable to the final model performance.

\begin{table}[!t]
\centering
\begin{tabular}{ l | c c c }
\toprule
 \backslashbox{Method}{Metric} & \textbf{OP} & \textbf{OR} &\textbf{OF1} \\ \midrule
Baseline & 80.01 & 79.43 & 78.73 \\ 
Ours - ResNet Indep & 82.77 & 82.65 &  82.65 \\ 
Ours - No MP & 81.66 & 82.57 & 81.47 \\ 
\textbf{Ours - final} & \textbf{83.53} & \textbf{84.16} & \textbf{83.35} \\
\toprule
\end{tabular}
\caption{Quantitative results for Category level.}
\label{tab:results_cat_2}
\end{table}

\subsubsection*{Sub-category level}

For sub-category and attributes level, we can only compare the baseline for some categories, as the baseline method is, for each sub-category and attribute type, an instantiation of a pipeline of specific models, thus not covering all classes. Only our final proposed model and its variants have full coverage and only for those we can report global performance results.
Hence, for a fair comparison, we report performance results for specific categories (Dresses and Coats) in Table~\ref{tab:results_cat_3_dresses/coats}, and global results (over all categories) in Table~\ref{tab:results_cat_3_all}. 
The results in Table~\ref{tab:results_cat_3_dresses/coats} were obtained by selecting from the dataset the products belonging to either the Dresses or Coats categories. 

We claim that this comparison benefits the baseline model for the following reasons:
\begin{itemize}
\item In the baseline approach, a product is only fed into the sub-category model specialised for dresses if the generic category model had predicted the respective product to be a dress. It means that if the latter makes a wrong prediction, all the subsequent predictions (sub-category and attributes) will be automatically wrong. In our analysis, we assume that the generic category model is 100\% accurate;
\item Each specialised sub-category model has an output dimension space much smaller than our unified approach. Namely, for dresses, there are only 4 possible sub-categories, while our approach can theoretically predict any of the 95 sub-categories referred in Section~\ref{sec:Dataset}.
\end{itemize}

Analysing Table~\ref{tab:results_cat_3_all}, although our final method does not evaluate the best for all metrics, we see a result consistent with the ones obtained for category level (in Table~\ref{tab:results_cat_2}) regarding the importance of the message passing scheme (again, the model without this scheme - \emph{Ours - No MP} is the worst performing among the variants).

\begin{table}[!t]
\centering
\begin{tabular}{ l | c c c }
\toprule
\multicolumn{2}{c}{}
 & \textbf{Dresses | Coats} \\ 
 \backslashbox{Method}{Metric} & \textbf{OP} & \textbf{OR} &\textbf{OF1} \\  \midrule
Baseline & 58.24|\textbf{60.88} & \textbf{59.61}|\textbf{58.7} & 56.00|\textbf{54.43} \\ 
Ours - ResNet Indep & 56.2|42.98 & 54.76|44.93 &  54.73|40.58 \\ 
Ours - No MP & 57.46|48.89 & 48.73|44.03 & 51.56|43.35 \\ 
\textbf{Ours - final} & \textbf{59.41}|52.95 & 57.68|51.11 & \textbf{56.48}|49.49 \\
\toprule
\end{tabular}
\caption{Quantitative results for sub-category level for Dresses and Coats categories.}
\label{tab:results_cat_3_dresses/coats}
\end{table}

\begin{table}[!t]
\centering
\begin{tabular}{ l | c c c }
\toprule
 \backslashbox{Method}{Metric} & \textbf{OP} & \textbf{OR} &\textbf{OF1} \\ \midrule
Ours - ResNet Indep & \textbf{45.74} & 34.90 &  \textbf{29.60} \\ 
Ours - No MP & 42.03 & 34.21 & 29.20 \\ 
\textbf{Ours - final} & 42.68 & \textbf{37.00} & 29.39 \\
\toprule
\end{tabular}
\caption{Quantitative results for sub-category level over all categories.}
\label{tab:results_cat_3_all}
\end{table}

\subsubsection*{Attribute level}

As discussed above, and similarly to the sub-category level, we show separate results for specific categories in Table~\ref{tab:results_att_dresses/coats} and global results in Table~\ref{tab:results_att_all}. Note that for the same reasons as in the sub-category level, the baseline approach has some advantage under this type of comparison.
Considering the examined multi-label metrics, and differently from the sub-category results for the specific categories, our proposed method beats the baseline for both Dresses and Coats for all but one metric (OP). Importantly, the F1$@$K improves over the OF1 (that considers all predicted attributes), showing that the meaningful attributes are indeed predicted with high scores. 
Furthermore, our final model outperforms its two variants for all metrics except one for each category, supporting once more our design choices.

\begin{table*}[!t]
\centering
\begin{tabular}{ l | c c c c c c c}
\toprule
\multicolumn{4}{c}{}
 & \textbf{Dresses | Coats} \\ 
 \backslashbox{Method}{Metric} & \textbf{OP} & \textbf{OR} &\textbf{OF1} & \textbf{P$@$k}  & \textbf{R$@$k} & \textbf{F1$@$k} & \textbf{AP} \\ \midrule

Baseline &  \textbf{59.37}| \textbf{50.80} & 35.48|28.9 & 38.04|23.05 & 55.10|49.64 & 45.79|30.24 & 45.07|24.61 & 54.26|50.81 \\ 
Ours - ResNet Indep & 43.51|45.43 & 85.66|77.95 & 55.6|55.01 & 60.92|60.48 & 57.97|57.28 &56.74|55.14 & 55.99|48.38 \\ 
Ours - No MP & 44.22|44.05 & 82.52|70.84 & 54.92|52.58 & 59.15|55.49 & 57.63|\textbf{59.11} & 55.82|52.27 & 53.82|47.78\\ 
\textbf{Ours - final} & 46.10|43.86 & \textbf{86.00}|\textbf{80.23} & \textbf{58.61}|\textbf{55.26} & \textbf{65.34}|\textbf{63.75} & \textbf{65.37}|58.50 & \textbf{64.49}|\textbf{56.10} & \textbf{60.88}|\textbf{51.22}\\
\toprule
\end{tabular}
\caption{Quantitative results for Dresses and Coats attributes.}
\label{tab:results_att_dresses/coats}
\end{table*}

\begin{table*}[!t]
\centering
\begin{tabular}{ l | c c c c c c c}
\toprule
 \backslashbox{Method}{Metric} & \textbf{OP} & \textbf{OR} &\textbf{OF1} & \textbf{P$@$k}  & \textbf{R$@$k} & \textbf{F1$@$k} & \textbf{AP} \\ \midrule
Ours - ResNet Indep & 47.55 & 85.16 & 58.51 & 66.63 & 66.33 & 64.63 & 59.04 \\
Ours - No MP & 47.17 & 84.51 & 58.04 & 65.25 & 66.09 & 63.40 & 57.79 \\
\textbf{Ours - final} & \textbf{49.22} & \textbf{86.75} & \textbf{60.60} & \textbf{69.19} & \textbf{70.78} & \textbf{68.86} & \textbf{61.91} \\
\toprule
\end{tabular}
\caption{Quantitative results for attributes for all classes.}
\label{tab:results_att_all}
\end{table*}

\begin{figure*}[!t]
\includegraphics[width=0.9\textwidth]{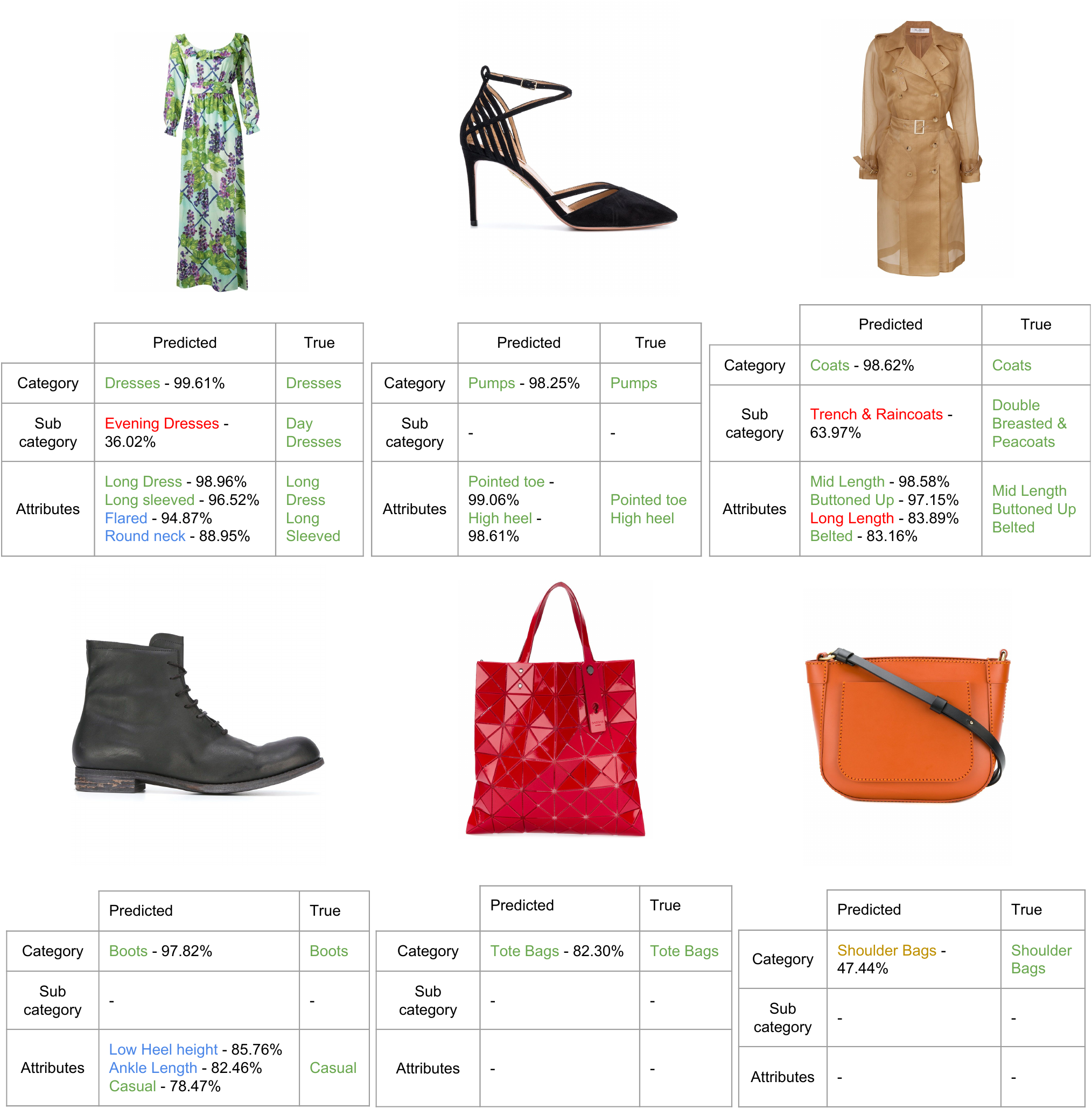}
\caption{Examples of prediction results on some products from our dataset. We compare the predictions with the ground truth annotations. Correct predictions (matching the ground-truth) are shown in green, incorrect are in red, correct predictions missing in the ground truth annotations are in blue, and correct predictions but with a low confidence are in yellow.}
\label{fig:qual_results}
\end{figure*}

Another interesting result noted from our evaluation is that the average number of attributes estimated per product by our proposed model is 2.07, much higher than the average number of manual attributes 0.96.
Thus, even if some attributes are wrongly predicted, it seems safe to say that the model, by inferring missing entries, tries to generalise; that is, it can estimate attributes that were missing in manual labelling, thus providing a more complete (and probably more consistent) description across products. We highlight that we do not directly address the missing data issue, however our proposed model seems to de-emphasize its harmful effect, as it is able to generalize attributes within similar products. This behaviour can be observed on some products of Figure~\ref{fig:qual_results}, namely for the dress and the boot, where the attributes shown in blue were not manually assigned in the ground truth but match the visual characteristics of the products, thus extending the captured visual properties.

Additionally, as our proposed model employs a unified design, that considers (differently from the baseline) categories, sub-categories and attributes altogether, the issue of associating inconsistent sub-categories or attributes to categories arises. To test and discard this possibility we listed the co-occurrences of categories and sub-categories as well as of categories and attributes and found that less than 1\% of these pairs were inconsistent (i.e., a skirt sub-category or attribute is assigned to a top category). This result points out that the model is, in fact, able to correctly capture the relations among concepts across levels. 

Some qualitative results of our proposed method (predictions for different types of products) can be visualised in Figure~\ref{fig:qual_results}. 

From the previous results we can pinpoint some limitations of our method. Stemming from the fact that we only use a single front-side image for each product there are some ambiguities such as scale (seen in Figure~\ref{fig:qual_results} for the coat with mid \emph{versus} long length attribute prediction, and for the last bag with the low confidence score obtained for shoulder bag category prediction), or details that are not visible from the front side.
The majority of these limitations, however, could be addressed by adopting a multi-view approach, where several images (views) of the same product are used to train the model. 

\section{Conclusion}
\label{sec:conclusion}

In this paper, we present a novel unified approach to categorise and predict attributes of fashion products. Our approach relies on the principle work of~\citep{Hu2016} by jointly learning different concept levels from a hierarchy tree, thus exploring the relations among labels. Our experimental analysis shows improvements for all categorisation levels upon a set of models specialised for each concept level based on the state-of-the-art (our baseline), in addition to allowing a full coverage of all product types. Moreover, the two variants of our final model seem to validate our design choices (sharing the final ResNet layers and including the adapted message passing scheme from~\citep{Hu2016}). 
Importantly, we have shown that with a single model (our unified approach) we achieve competitive results in sub-category classification and even outperform the framework of a series of specialised models for both category classification and attribute labelling. 
In particular, for the multi-label problem tackled for the attributes level, we have verified that our model is able to predict a higher number of attributes globally, thus producing more consistent and complete annotations over all types of products than the ground truth annotations. We believe these findings can bring substantial benefits to the problem of image labelling for fashion e-commerce.



\begin{acks}
  The authors would like to thank all team members of the search department from Farfetch. Their support allow us to deliver and assess the quality of our algorithms and the benefits provided to our customers. This work was partially funded by FCT via grant [PD/BD/114430/2016] and project [UID/EEA/50009/2013].
\end{acks}

\bibliographystyle{ACM-Reference-Format}
\bibliography{bibliography}


\begin{thebibliography}{22}


\ifx \showCODEN    \undefined \def \showCODEN     #1{\unskip}     \fi
\ifx \showDOI      \undefined \def \showDOI       #1{#1}\fi
\ifx \showISBNx    \undefined \def \showISBNx     #1{\unskip}     \fi
\ifx \showISBNxiii \undefined \def \showISBNxiii  #1{\unskip}     \fi
\ifx \showISSN     \undefined \def \showISSN      #1{\unskip}     \fi
\ifx \showLCCN     \undefined \def \showLCCN      #1{\unskip}     \fi
\ifx \shownote     \undefined \def \shownote      #1{#1}          \fi
\ifx \showarticletitle \undefined \def \showarticletitle #1{#1}   \fi
\ifx \showURL      \undefined \def \showURL       {\relax}        \fi
\providecommand\bibfield[2]{#2}
\providecommand\bibinfo[2]{#2}
\providecommand\natexlab[1]{#1}
\providecommand\showeprint[2][]{arXiv:#2}

\bibitem[\protect\citeauthoryear{Chen, Gallagher, and Girod}{Chen
  et~al\mbox{.}}{2012}]%
        {ClothingAttDataset2012}
\bibfield{author}{\bibinfo{person}{Huizhong Chen}, \bibinfo{person}{Andrew
  Gallagher}, {and} \bibinfo{person}{Bernd Girod}.}
  \bibinfo{year}{2012}\natexlab{}.
\newblock \showarticletitle{{Describing Clothing by Semantic Attributes}}, In
  \bibinfo{booktitle}{IEEE European Conference on Computer Vision (ECCV),
  2012}.
\newblock \bibinfo{journal}{\emph{Lecture Notes in Computer Science}}
  \bibinfo{volume}{7574 LNCS}.
\newblock


\bibitem[\protect\citeauthoryear{Corbiere, Ben-Younes, Rame, and
  Ollion}{Corbiere et~al\mbox{.}}{2015}]%
        {Corbiere2017}
\bibfield{author}{\bibinfo{person}{Charles Corbiere}, \bibinfo{person}{Heidi
  Ben-Younes}, \bibinfo{person}{Alexandre Rame}, {and} \bibinfo{person}{Charles
  Ollion}.} \bibinfo{year}{2015}\natexlab{}.
\newblock \showarticletitle{{Leveraging Weakly Annotated Data for Fashion Image
  Retrieval and Label Prediction}}. In \bibinfo{booktitle}{\emph{Proceedings of
  the Workshop from IEEE International Conference on Computer Vision, (ICCV
  Workshop), 2017}}.
\newblock


\bibitem[\protect\citeauthoryear{Deng, Ding, Jia, Frome, Murphy, Bengio, Li,
  Neven, and Adam}{Deng et~al\mbox{.}}{2014}]%
        {Deng2014}
\bibfield{author}{\bibinfo{person}{Jia Deng}, \bibinfo{person}{Nan Ding},
  \bibinfo{person}{Yangqing Jia}, \bibinfo{person}{Andrea Frome},
  \bibinfo{person}{Kevin Murphy}, \bibinfo{person}{Samy Bengio},
  \bibinfo{person}{Yuan Li}, \bibinfo{person}{Hartmut Neven}, {and}
  \bibinfo{person}{Hartwig Adam}.} \bibinfo{year}{2014}\natexlab{}.
\newblock \showarticletitle{{Large-scale object classification using label
  relation graphs}}, In \bibinfo{booktitle}{IEEE European Conference on
  Computer Vision (ECCV), 2014}.
\newblock \bibinfo{journal}{\emph{Lecture Notes in Computer Science}}
  \bibinfo{volume}{8689 LNCS}.
\newblock
\showISBNx{9783319105895}
\showISSN{16113349}


\bibitem[\protect\citeauthoryear{Deng, Dong, Socher, Li, Li, and Fei-Fei}{Deng
  et~al\mbox{.}}{2009}]%
        {imagenet_cvpr09}
\bibfield{author}{\bibinfo{person}{J. Deng}, \bibinfo{person}{W. Dong},
  \bibinfo{person}{R. Socher}, \bibinfo{person}{L.~J. Li}, \bibinfo{person}{K.
  Li}, {and} \bibinfo{person}{L. Fei-Fei}.} \bibinfo{year}{2009}\natexlab{}.
\newblock \showarticletitle{{ImageNet: A Large-Scale Hierarchical Image
  Database}}. In \bibinfo{booktitle}{\emph{IEEE Conference on Computer Vision
  and Pattern Recognition (CVPR), 2009}}.
\newblock


\bibitem[\protect\citeauthoryear{He, Zhang, Ren, and Sun}{He
  et~al\mbox{.}}{2016}]%
        {He2016}
\bibfield{author}{\bibinfo{person}{Kaiming He}, \bibinfo{person}{Xiangyu
  Zhang}, \bibinfo{person}{Shaoqing Ren}, {and} \bibinfo{person}{Jian Sun}.}
  \bibinfo{year}{2016}\natexlab{}.
\newblock \showarticletitle{{Deep Residual Learning for Image Recognition}}. In
  \bibinfo{booktitle}{\emph{IEEE Conference on Computer Vision and Pattern
  Recognition (CVPR), 2016}}.
\newblock


\bibitem[\protect\citeauthoryear{Hinton, Srivastava, Krizhevsky, Sutskever, and
  Salakhutdinov}{Hinton et~al\mbox{.}}{2012}]%
        {Hinton_droput2012}
\bibfield{author}{\bibinfo{person}{Geoffrey~E. Hinton}, \bibinfo{person}{Nitish
  Srivastava}, \bibinfo{person}{Alex Krizhevsky}, \bibinfo{person}{Ilya
  Sutskever}, {and} \bibinfo{person}{Ruslan Salakhutdinov}.}
  \bibinfo{year}{2012}\natexlab{}.
\newblock \showarticletitle{{Improving neural networks by preventing
  co-adaptation of feature detectors}}.
\newblock \bibinfo{journal}{\emph{CoRR}}  \bibinfo{volume}{abs/1207.0580}.
\newblock
\urldef\tempurl%
\url{http://arxiv.org/abs/1207.0580}
\showURL{%
\tempurl}


\bibitem[\protect\citeauthoryear{Hu, Zhou, Deng, Liao, and Mori}{Hu
  et~al\mbox{.}}{2016}]%
        {Hu2016}
\bibfield{author}{\bibinfo{person}{Hexiang Hu}, \bibinfo{person}{Guang-Tong
  Zhou}, \bibinfo{person}{Zhiwei Deng}, \bibinfo{person}{Zicheng Liao}, {and}
  \bibinfo{person}{Greg Mori}.} \bibinfo{year}{2016}\natexlab{}.
\newblock \showarticletitle{{Learning Structured Inference Neural Networks with
  Label Relations}}. In \bibinfo{booktitle}{\emph{IEEE Conference on Computer
  Vision and Pattern Recognition (CVPR), 2016}}.
\newblock


\bibitem[\protect\citeauthoryear{Kingma and Ba}{Kingma and Ba}{2015}]%
        {Kingma2015}
\bibfield{author}{\bibinfo{person}{Diederik Kingma} {and}
  \bibinfo{person}{Jimmy Ba}.} \bibinfo{year}{2015}\natexlab{}.
\newblock \showarticletitle{{ADAM: A method for stochastic optimization}}. In
  \bibinfo{booktitle}{\emph{International Conference on Learning
  Representations (ICLR)}}.
\newblock


\bibitem[\protect\citeauthoryear{Krizhevsky, Sutskever, and Hinton}{Krizhevsky
  et~al\mbox{.}}{2012}]%
        {Krizhevsky2012}
\bibfield{author}{\bibinfo{person}{A. Krizhevsky}, \bibinfo{person}{I.
  Sutskever}, {and} \bibinfo{person}{G. Hinton}.}
  \bibinfo{year}{2012}\natexlab{}.
\newblock \showarticletitle{{ImageNet classification with deep convolutional
  neural networks}}. In \bibinfo{booktitle}{\emph{25th Conference on Neural
  Information Processing Systems (NIPS)}}.
\newblock


\bibitem[\protect\citeauthoryear{Laenen, Zoghbi, and Moens}{Laenen
  et~al\mbox{.}}{2017}]%
        {Laenen2017}
\bibfield{author}{\bibinfo{person}{Katrien Laenen}, \bibinfo{person}{Susana
  Zoghbi}, {and} \bibinfo{person}{Marie-Francine Moens}.}
  \bibinfo{year}{2017}\natexlab{}.
\newblock \showarticletitle{{Cross-modal Search for Fashion Attributes}}. In
  \bibinfo{booktitle}{\emph{Proceedings of the KDD 2017 Workshop on Machine
  Learning Meets Fashion}}. \bibinfo{publisher}{ACM}.
\newblock


\bibitem[\protect\citeauthoryear{Liu, Xiang, Hospedales, Yang, and Sun}{Liu
  et~al\mbox{.}}{2017}]%
        {Liu2017}
\bibfield{author}{\bibinfo{person}{Feng Liu}, \bibinfo{person}{Tao Xiang},
  \bibinfo{person}{Timothy~M. Hospedales}, \bibinfo{person}{Wankou Yang}, {and}
  \bibinfo{person}{Changyin Sun}.} \bibinfo{year}{2017}\natexlab{}.
\newblock \showarticletitle{{Semantic Regularisation for Recurrent Image
  Annotation}}. In \bibinfo{booktitle}{\emph{IEEE Conference on Computer Vision
  and Pattern Recognition (CVPR), 2017}}.
\newblock


\bibitem[\protect\citeauthoryear{Liu, Luo, Qiu, Wang, and Tang}{Liu
  et~al\mbox{.}}{2016}]%
        {LiuDeepFashion2016}
\bibfield{author}{\bibinfo{person}{Ziwei Liu}, \bibinfo{person}{Ping Luo},
  \bibinfo{person}{Shi Qiu}, \bibinfo{person}{Xiaogang Wang}, {and}
  \bibinfo{person}{Xiaoou Tang}.} \bibinfo{year}{2016}\natexlab{}.
\newblock \showarticletitle{{DeepFashion: Powering Robust Clothes Recognition
  and Retrieval with Rich Annotations}}. In \bibinfo{booktitle}{\emph{IEEE
  Conference on Computer Vision and Pattern Recognition (CVPR), 2016}}.
\newblock


\bibitem[\protect\citeauthoryear{Long, Cao, Wang, and Yu}{Long
  et~al\mbox{.}}{2017}]%
        {Long2017}
\bibfield{author}{\bibinfo{person}{Mingsheng Long}, \bibinfo{person}{Zhangjie
  Cao}, \bibinfo{person}{Jianmin Wang}, {and} \bibinfo{person}{Philip~S. Yu}.}
  \bibinfo{year}{2017}\natexlab{}.
\newblock \showarticletitle{{Learning Multiple Tasks with Multilinear
  Relationship Networks}}. In \bibinfo{booktitle}{\emph{31st Conference on
  Neural Information Processing Systems (NIPS)}}.
\newblock


\bibitem[\protect\citeauthoryear{Niu, Lu, Wen, Xiang, and Chang}{Niu
  et~al\mbox{.}}{2017}]%
        {Niu2017}
\bibfield{author}{\bibinfo{person}{Yulei Niu}, \bibinfo{person}{Zhiwu Lu},
  \bibinfo{person}{Ji-Rong Wen}, \bibinfo{person}{Tao Xiang}, {and}
  \bibinfo{person}{Shih-Fu Chang}.} \bibinfo{year}{2017}\natexlab{}.
\newblock \showarticletitle{{Multi-Modal Multi-Scale Deep Learning for
  Large-Scale Image Annotation}}.
\newblock \bibinfo{journal}{\emph{CoRR}}  \bibinfo{volume}{abs/1709.01220}.
\newblock
\urldef\tempurl%
\url{http://arxiv.org/abs/1709.01220}
\showURL{%
\tempurl}


\bibitem[\protect\citeauthoryear{Reddi, Kale, and Kumar}{Reddi
  et~al\mbox{.}}{2018}]%
        {Reddi2018}
\bibfield{author}{\bibinfo{person}{Sashank~J. Reddi}, \bibinfo{person}{Satyen
  Kale}, {and} \bibinfo{person}{Sanjiv Kumar}.}
  \bibinfo{year}{2018}\natexlab{}.
\newblock \showarticletitle{{On the Convergence of Adam and Beyond}}. In
  \bibinfo{booktitle}{\emph{International Conference on Learning
  Representations (ICLR)}}.
\newblock


\bibitem[\protect\citeauthoryear{Simard, Steinkraus, and Platt}{Simard
  et~al\mbox{.}}{2003}]%
        {Simard2003}
\bibfield{author}{\bibinfo{person}{P. Simard}, \bibinfo{person}{D. Steinkraus},
  {and} \bibinfo{person}{J.~C. Platt}.} \bibinfo{year}{2003}\natexlab{}.
\newblock \showarticletitle{{Best practices for convolutional neural networks
  applied to visual document analysis}}. In \bibinfo{booktitle}{\emph{Proc.
  Int. Conf. Document Anal. Recognit.}}
\newblock


\bibitem[\protect\citeauthoryear{Simonyan and Zisserman}{Simonyan and
  Zisserman}{2014}]%
        {Simonyan2014}
\bibfield{author}{\bibinfo{person}{Karen Simonyan} {and}
  \bibinfo{person}{Andrew Zisserman}.} \bibinfo{year}{2014}\natexlab{}.
\newblock \showarticletitle{{Very Deep Convolutional Networks for Large-Scale
  Image Recognition}}.
\newblock \bibinfo{journal}{\emph{CoRR}}  \bibinfo{volume}{abs/1409.1556}.
\newblock
\urldef\tempurl%
\url{http://arxiv.org/abs/1409.1556}
\showURL{%
\tempurl}


\bibitem[\protect\citeauthoryear{Sun, Wu, Chen, and Peng}{Sun
  et~al\mbox{.}}{2015}]%
        {Sun2015}
\bibfield{author}{\bibinfo{person}{Guang-Lu Sun}, \bibinfo{person}{Xiao Wu},
  \bibinfo{person}{Hong-Han Chen}, {and} \bibinfo{person}{Qiang Peng}.}
  \bibinfo{year}{2015}\natexlab{}.
\newblock \showarticletitle{{Clothing Style Recognition using Fashion Attribute
  Detection}}. In \bibinfo{booktitle}{\emph{Proceedings of the 8th
  International Conference on Mobile Multimedia Communications}}.
\newblock


\bibitem[\protect\citeauthoryear{Szegedy, Liu, Jia, Sermanet, and Reed}{Szegedy
  et~al\mbox{.}}{2015}]%
        {Szegedy2016}
\bibfield{author}{\bibinfo{person}{Christian Szegedy}, \bibinfo{person}{Wei
  Liu}, \bibinfo{person}{Yangqing Jia}, \bibinfo{person}{Pierre Sermanet},
  {and} \bibinfo{person}{Scott Reed}.} \bibinfo{year}{2015}\natexlab{}.
\newblock \showarticletitle{{Going Deeper with Convolutions}}. In
  \bibinfo{booktitle}{\emph{IEEE Conference on Computer Vision and Pattern
  Recognition (CVPR), 2015}}.
\newblock


\bibitem[\protect\citeauthoryear{Wang, Yang, Mao, Huang, Huang, and Xu}{Wang
  et~al\mbox{.}}{2016}]%
        {Wang2016}
\bibfield{author}{\bibinfo{person}{Jiang Wang}, \bibinfo{person}{Yi Yang},
  \bibinfo{person}{Junhua Mao}, \bibinfo{person}{Zhiheng Huang},
  \bibinfo{person}{Chang Huang}, {and} \bibinfo{person}{Wei Xu}.}
  \bibinfo{year}{2016}\natexlab{}.
\newblock \showarticletitle{{CNN-RNN: A Unified Framework for Multi-label Image
  Classification}}. In \bibinfo{booktitle}{\emph{IEEE Conference on Computer
  Vision and Pattern Recognition (CVPR), 2016}}.
\newblock


\bibitem[\protect\citeauthoryear{Yamaguchi, Kiapour, Ortiz, and Berg}{Yamaguchi
  et~al\mbox{.}}{2012}]%
        {Fashionista2012}
\bibfield{author}{\bibinfo{person}{Kota Yamaguchi}, \bibinfo{person}{{M. Hadi}
  Kiapour}, \bibinfo{person}{{Luis E.} Ortiz}, {and} \bibinfo{person}{{Tamara
  L.} Berg}.} \bibinfo{year}{2012}\natexlab{}.
\newblock \showarticletitle{Parsing clothing in fashion photographs}. In
  \bibinfo{booktitle}{\emph{IEEE Conference on Computer Vision and Pattern
  Recognition (CVPR), 2012}}.
\newblock


\bibitem[\protect\citeauthoryear{Yang and Ramanan}{Yang and Ramanan}{2015}]%
        {Yang2015}
\bibfield{author}{\bibinfo{person}{Songfan Yang} {and} \bibinfo{person}{Deva
  Ramanan}.} \bibinfo{year}{2015}\natexlab{}.
\newblock \showarticletitle{{Multi-scale recognition with DAG-CNNs}}. In
  \bibinfo{booktitle}{\emph{Proceedings of the IEEE International Conference on
  Computer Vision, (ICCV), 2015}}.
\newblock


\end{thebibliography}

\end{document}